\documentclass{article} 
\usepackage{nips13submit_e,times}
\usepackage{hyperref}
\usepackage{url}
\usepackage{graphicx}
\usepackage{mathtools}

\title{Conditioning LSTM Decoder and Bi-directional Attention Based Question Answering System}

\author{
Heguang Liu\thanks{ Senior Software Engineer at Uber Technologies, Inc and SCPD student at Stanford University.} \\
Uber Technologies, Inc
\\Stanford University\\
\texttt{heguangl@uber.com} \\
}

\nipsfinalcopy 

\begin{document}

\maketitle

\begin{abstract}
Applying neural-networks on Question Answering has gained increasing popularity in recent years. In this paper, I implemented a model with Bi-directional attention flow layer, connected with a Multi-layer LSTM encoder, connected with one start-index decoder and one conditioning end-index decoder. I introduce a \textbf{new end-index decoder layer, conditioning on start-index output}. Experiment shows this has increased model performance by \textbf{15.16\%}. For prediction, I proposed a new  \textbf{smart-span equation}, rewarding both short answer length and high probability in start-index and end-index, which further improved the prediction accuracy. The best single model achieves an F1 score of \textbf{73.97\%} and EM score of \textbf{64.95\%} on test set.
\end{abstract}

\section{Introduction}
Deep Neural Networks have gained significant popularity recently and have been applied to many Natural Language Processing (NLP) systems, such as machine translation, speech recognition, sentiment analysis, document summarization and reading comprehension. Among these applications, question answering, which is used in chatbots and dialogue agents, has attracted many attentions. The goal of question answering system is to provide an accurate answer to a given question based on a given context paragraph. This requires complex model interactive between question and context. 

In this paper, I explored a variety of improvements of SQuAD approaches, implement a question answering system using a Bi-direction attention flow layer, connected with a Multi-layer LSTM encoder and decoder for start-index. On top of this, I introduce a new end-index decoder layer, conditioning on start-index output. Experiment shows this has increased model performance by 15.16\%. For prediction, I introduce a new smart-span equation, which rewarding both short answer length and high probability in start and end-index, which further improved the prediction accuracy. 

\section{Background}
Stanford Question Answering Dataset(SQuAD)[1] (Rajpurkar et al., 2016) is a reading comprehension dataset, consisting of 100,000+ question-answer pairs on 500+ articles. SQuAD has led to many research papers and significant breakthroughs in building effective reading comprehension systems. One big breakthrough is Bi-Directional Attention Flow[2] (Seo et al., 2016), which enables the system to focus on the ROI in the context paragraph that is most relevant to the question, as well as the ROI in the question that is most relevant to the context. Another trending direction is Conditioning Prediction. "Match-LSTM and Answer Pointer"[3](Wang et al., 2016) proposed a Pointer Network that outputs a probability distribution over locations in the context. DrQA(Chen et al., 2017) explored a test-time smarter span selection approach, based on the fact answer ofter consists less than 15 words. 

\section{Approach}
\subsection{Architecture}
For this project, I propose the following 6-layer model architecture, as showed in Figure 1. 
\begin{figure}[h]
\includegraphics[width=\linewidth]{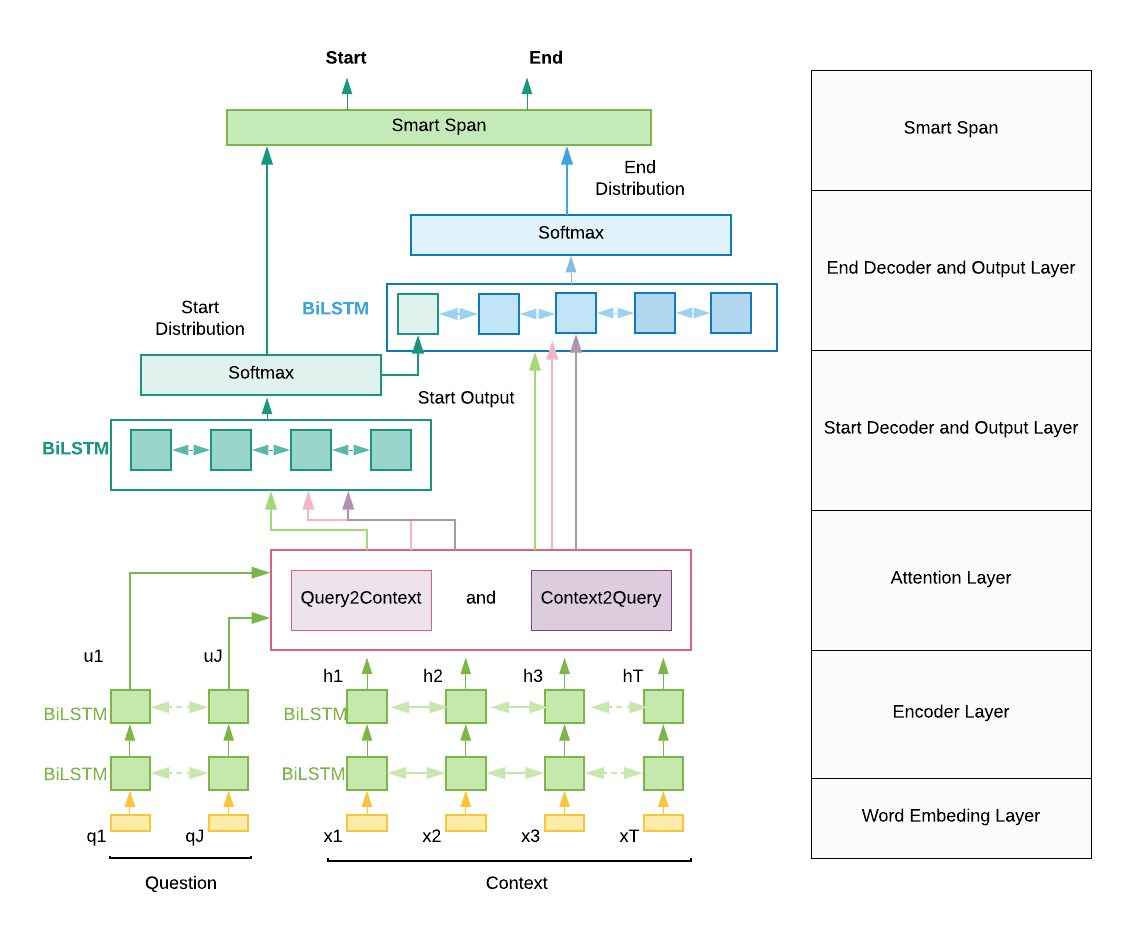}
\caption{Conditioning LSTM Decoder and Bi-directional Attention Based Question Answering System Architecture}
\end{figure}
\subsubsection{Embedding Layer}
Embedding layer maps each word in question and context to a vector space. For this project, I uses pre-trained Stanford GloVe [5] embeddings, during training, the embeddings are constants. My experiment shows that, increasing embedding dimension from 100 to 300 doesn't provide much model improvements, but it will make the training 40\% slower and more prone to over-fitting. So I use GloVe 100-dimension embeddings for all following models.
\subsubsection{Encoder Layer}
The context and question embedding sequence are feed into a 2-Layer Bi-directional LSTM to encode to forward and backward hidden states.
\subsubsection{Attention Layer}
To enable the model focus on the part of the question that is relevant to the context as well as part of the context that is relevant to the question, I use Bidirectional Attention Flow Layer described at [2].

\subsubsection{Start Decoder and End-Decoder}
For this project, I introduced two Bi-LSTM decoders to predict start-index and end-index position distribution separately. The end-index decoder will take start-index decoder output combined with post-attention layer hidden states as the input so that the end-index prediction is conditioning on the start-index distribution. Compared with baseline model, this approach adds a Bi-LSTM model layer and the two models are no longer sharing weights due to they are different problems by nature. This changes alone has magically increased the F1 score by 13\% when comparing with baseline model. 
\subsubsection{Output Layer}
Output Layer is a fully connected layer followed by a ReLU non-linearity. Finally, we apply softmax function to start-index logits and conditioning end-index logits to generate the probability distribution for both start and end position. Loss function is the sum of the cross-entropy loss of start and end indexes.
\subsubsection{Smart Span}
After analyzing failed session, I noticed that if a shorter answer and a longer answer has similar $p_{start}*p_{end}$ probability, the ground-truth answer is often the shorter answer. To capture this information, I introduced an smart-span equation to measure start-end pair-wise probability. This function is rewarding short answer length as well as high start-end-index probability at the same time.
\textbf{$$p(start, end)=\frac{p_{start}*p_{end}}{log(end-start+1)+1}$$}

\section{Experiments}
\subsection{Training Data}
SQuAD[1] provides word sequences of paragraphs sourced from 500+ Wikipedia articles and word sequence of 100K questions and answers, crowdsourced using Amazon Mechanical Turk. The answer are always taken directly from the paragraph. To minimize the noise in human-curated ground-truth answer, SQuAD provides three human answers for each question, and official evaluation will take the highest score among the three answers. The data is split into 80\% training and 10\% development set and 10\% withhold test set. 

\par Based on the analysis on 86,318 training data, \textbf{98.98\%} of answers is less than 20 words. \textbf{98.34\%} of context is less than 300 words.
\begin{figure}[h]
\begin{tabular}{cc}
  \includegraphics[width=65mm]{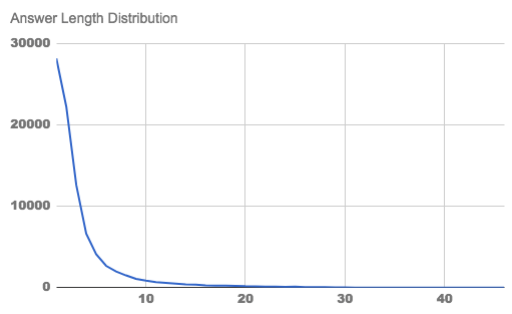} &   \includegraphics[width=65mm]{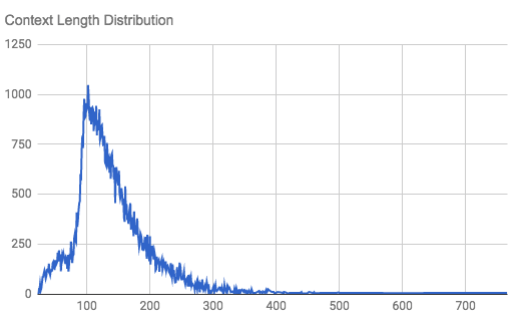} \\
(a) Answer Length Distribution & (b) Context Length Distribution  \\[6pt]
\end{tabular}
\caption{Training and Dev Loss Tensorboard}
\end{figure}

\par In this paper, I use pre-trained word embedding GloVe[5]. GloVe is trained on the 6 Billions words of Wikipedia and Gigaword. It has choice of dimensionality 50/100/200/300, each contains 400k lowercase words.

\subsection{Evaluation Metrics}
Model Performance is measured via F1 and Exact Match(EM) score.
\par F1 is a harmonic mean of precision and recall. This is a less strict metrics.
$$F1=\frac{2*Precision*Recall}{Precision+Recall}$$
\par EM is a binary measure of whether the system output matches the ground truth answer exactly. This is a fairly strict metrics.
$$EM=\frac{Exact Match Count}{Total Count}$$

\subsection{Training Details}
The code is implemented using Tensorflow v1.4 [6] framework, and written in python. All the models are trained on GPU of Microsoft Azure NV12 standard VM, which has 2 Nvidia Tesla 60M GPU pre-installed. The most complex model takes 2.5 days to converge.
\par I start the training with a baseline model. After analyzing, I found Dev set converge at F1=42\% and the training F1 is only ~70\%. To make the model complex enough to capture the training information, 2-Layer Bi-direction LSTMs and BiDAF is introduced to improve the F1 score to 53.59\%. After analyzing failed predictions, I noticed some answers are abnormally long. To capture this information, instead of predict end-index position independently, its prediction can be conditioning on start-index output. This final approach achieves 15.16\% increase in F1 score. 

\begin{figure}[h]
\begin{tabular}{cc}
  \includegraphics[width=65mm]{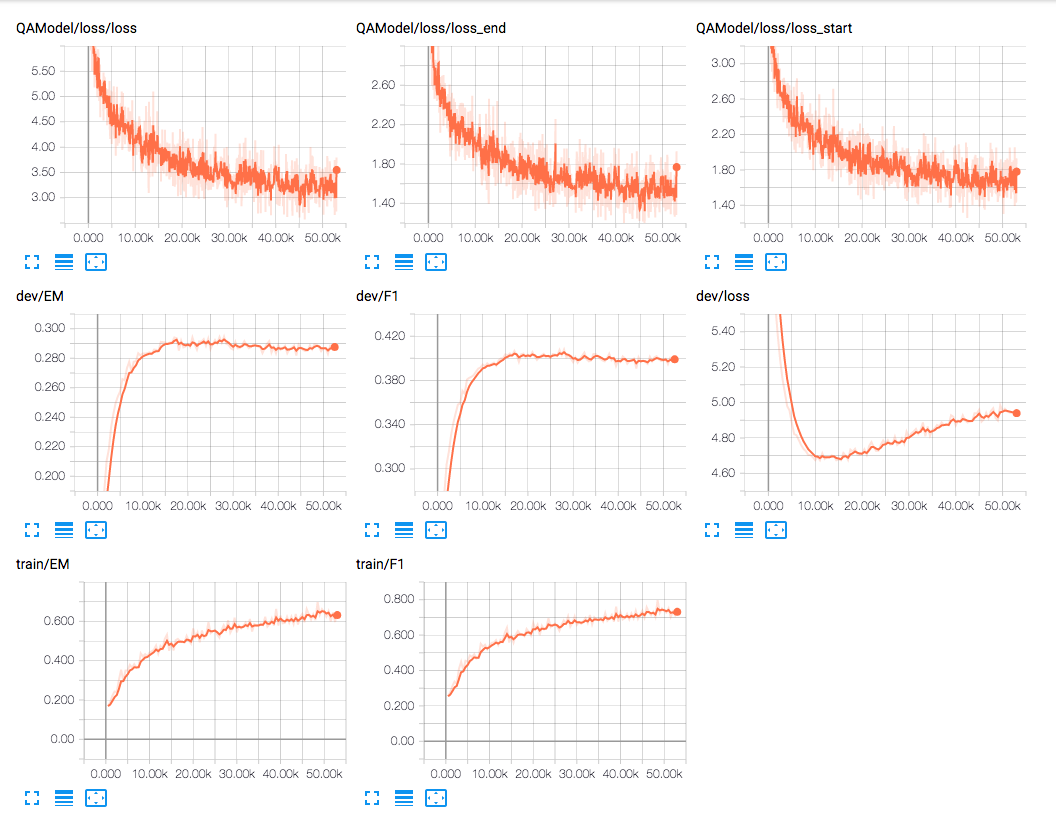} &   \includegraphics[width=65mm]{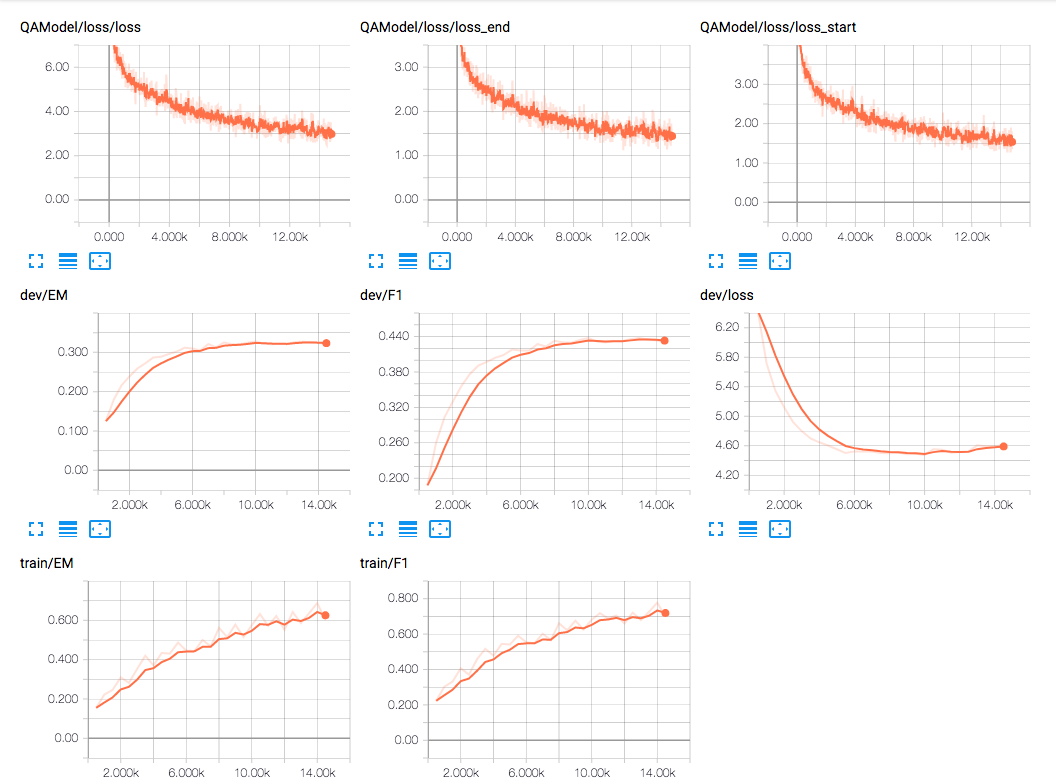} \\
(a) Baseline Model & (b) 2-Layer Bi-LSTMs \\[6pt]
 \includegraphics[width=65mm]{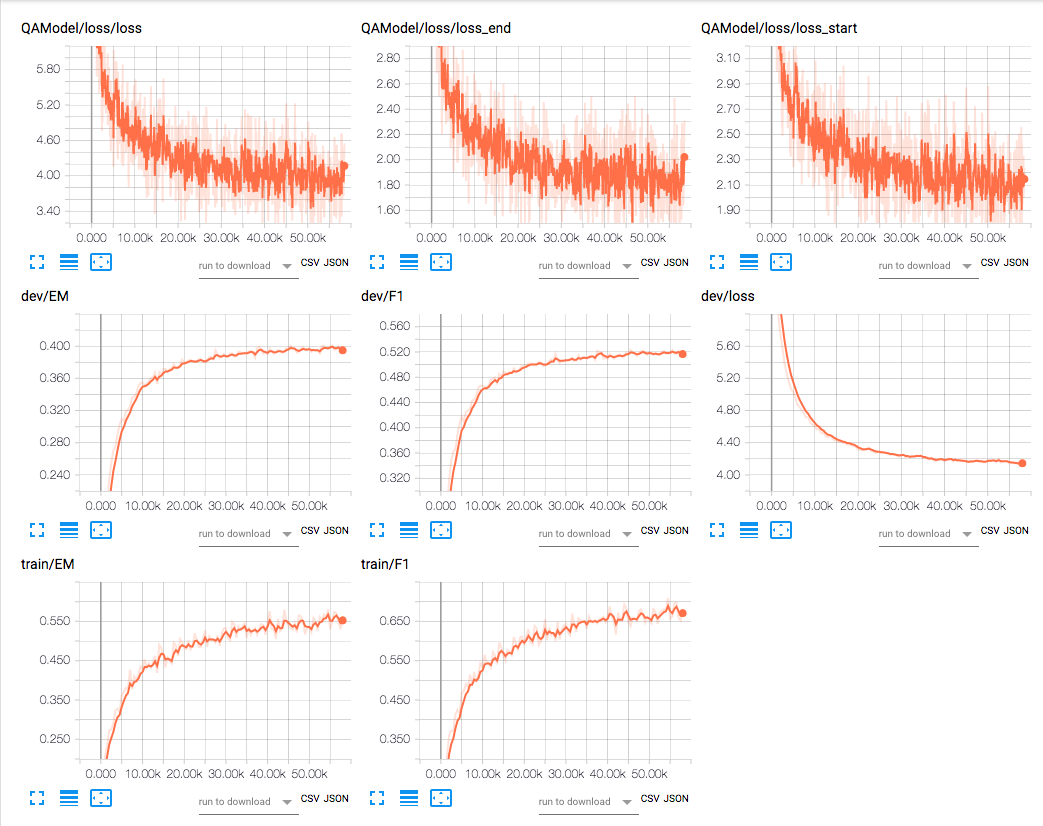} &   \includegraphics[width=65mm]{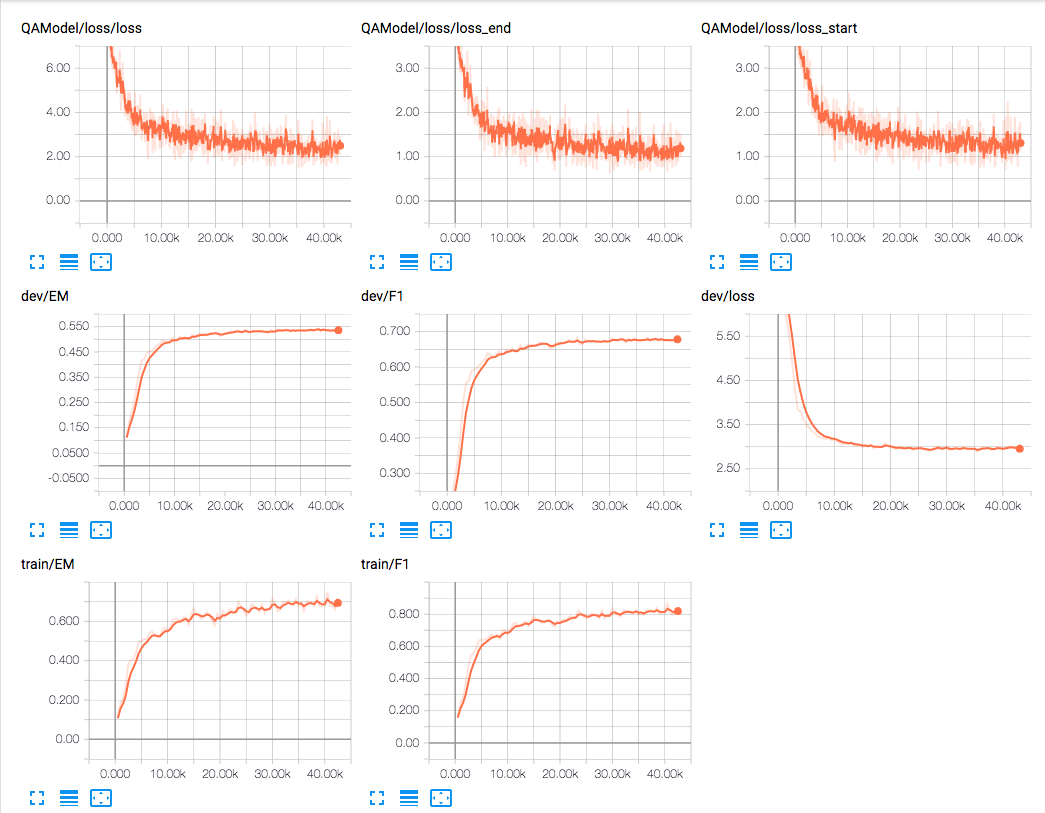} \\
(c) 2-Layer Bi-LSTMs with BiDAF & (d) Bi-LSTMs, BiDAF and Conditioning Decoders  \\[6pt]
\end{tabular}
\caption{Training and Dev Loss Tensorboard}
\end{figure}

\begin{table}[h]
\caption{Model Comparison}
\begin{center}
\begin{tabular}{llllll}
\multicolumn{1}{c}{\bf Model} &
\multicolumn{1}{c}{\bf F1 } &
\multicolumn{1}{c}{\bf EM } &
\multicolumn{1}{c}{\bf Parameter Size} &
\multicolumn{1}{c}{\bf Training Time} &
\multicolumn{1}{c}{\bf Batch Size} 
\\ 
\hline \\
Baseline        & 44.05 & 34.53  & 521,802 & 1.2s & 100 \\
2-Layer BiLSTMs     &53.59 & 44.63  & 1,283,802 & 2.0s & 100 \\
2-Layer BiLSTMs with BiDAF  & 58.16 & 48.76 & 2,405,103 & 2.7s  & 100 \\
2-Layer BiLSTMs with BiDAF, ConDecoders & 73.82 & 64.54  & 3,048,603 & 2.8s & 40 \\
\end{tabular}
\end{center}
\end{table}

\subsection{Model Parameters Tuning}
On top of the 2-Layer BiLSTMs with BiDAF, Conditioning Decoders model architecture, we start model hyper parameters tuning. Because most of the context are less than 300 words. So I truncate the context words after 300 words, to improve training speed. For this project, the model tuning is mainly focused on hidden size, dropout rate and embedding size.  The converged tuning result after 50,000 iterations is list in Table 3.
\begin{table}[h]
\caption{Model Parameters Tuning}
\begin{center}
\begin{tabular}{llllll}
\multicolumn{1}{c}{\bf Model} &
\multicolumn{1}{c}{\bf F1 } &
\multicolumn{1}{c}{\bf EM } &
\multicolumn{1}{c}{\bf Hidden Size} &
\multicolumn{1}{c}{\bf Dropout Rate} &
\multicolumn{1}{c}{\bf Embedding Size} 
\\ 
\hline \\
Model 1        & 73.83 & 64.54  & 150 & 0.20 & 100 \\
Model 2        & 71.68 & 61.97  & 150 & 0.25 & 200 \\
Model 3        & 72.66 & 62.16 & 200 & 0.30  & 100 \\
Model 4        & 73.32 & 63.59  & 250 & 0.20 & 100 \\
Model 5        & 73.73 & 64.51  & 150 & 0.15 & 100 \\
\end{tabular}
\end{center}
\end{table}

\par Table 3 shows that
\begin{itemize}
\item When $HiddenSize = 150$, $Dropout = 0.2$ is a good choice. Increase will lead to high variance model and decrease will lead to over-fitting in early stage
\item Increase $HiddenSize$ from 150 to 250 doesn't improve the performance much. 
\item $EmbeddingSize = 100$ has a reasonable good performance. Increase embedding size from 100 to 200 doesn't affect model performance much.
\end{itemize}

\section{Result and Analysis}
\subsection{Category Analysis}
To better understand the result, I did an category analysis on the 10,391 Dev predictions. The result per category analysis in Table 4 shows that questions contain "When" and "Who" are clearly better than average, while questions contain "Why" are performing below average. 
\begin{table}[h]
\caption{Evaluation by Question Category}
\begin{center}
\begin{tabular}{lll}
\multicolumn{1}{c}{\bf Category} &
\multicolumn{1}{c}{\bf F1 } &
\multicolumn{1}{c}{\bf EM } 
\\ 
\hline \\
Total        & 73.83 & 64.54    \\
Who        & 76.37 & 70.52  \\
When        & 83.07 & 77.50  \\
Where        & 73.32 & 63.77  \\
Why        & 65.71 & 41.44  \\
What        & 72.14 & 62.00  \\
Which       & 74.36 & 65.66  \\
How       & 74.24 & 65.15  \\
\end{tabular}
\end{center}
\end{table}

\subsection{Error Analysis}
After analyzing the model prediction on tiny-dev dataset, I find there are four common errors in my model prediction. 
\subsubsection{Wrong Position of Attention in Inverted Sentence}
\textbf{Context}: A 16-yard reception by Devin Funchess and a 12-yard run by Stewart then set up Gano's 39-yard field goal
\par \textbf{Question}: Who had a 12-yard rush on this drive?
\par \textbf{Answer}: Stewart
\par \textbf{Prediction}: Devin Funchess
\par \textbf{Reason and Proposal}: In this case, the order of subject and verb are reversed. Having part-of-speech tagging feature will help with the inverted sentences.
\subsubsection{Mis-understand Number and Special Characters}
\textbf{Context}: while Jonathan Stewart finished the drive with a 1-yard touchdown run, cutting the score to 10-7 with 11:28 left in the second quarter.
\par \textbf{Question}: How much time was left in the quarter when Stewart got the touchdown?
\par \textbf{Answer}: 11:28
\par \textbf{Prediction}: 10-7
\par \textbf{Reason and Proposal}: Word embedding doesn't capture the meaning of numbers and special characters. We can either introduce a number/special character embedding or introduce number format feature to help improve this case. 
\subsubsection{Incorrect Entity Detection}
\textbf{Context}: As opposed to broadcasts of primetime series, CBS broadcast special episodes of its late night talk shows as its lead-out programs for Super Bowl 50, beginning with a special episode of The Late Show with Stephen Colbert following the game.
\par \textbf{Question}: Which late night comedy host show played immediately after Super Bowl 50 ended?
\par \textbf{Answer}: The Late Show with Stephen Colbert
\par \textbf{Prediction}: stephen colbert
\par \textbf{Reason and Proposal}: The model falsely recognize stephen colbert as the name of a show. Include name entity feature will improve this issue.
\subsubsection{Inaccurate boundary}
\textbf{Context}: The Super Bowl 50 Host Committee has vowed to be "the most giving Super Bowl ever", and will dedicate 25 percent of all money it raises for philanthropic causes in the Bay Area.
\par \textbf{Question}: The Super Bowl 50 Host Committee said it would be the most what ever?
\par \textbf{Answer}: the most giving Super Bowl ever
\par \textbf{Prediction}: giving super bowl ever
\par \textbf{Reason and Proposal}: Due to the pre-processing tokenization logic, we loss the double quote information. We can tokenize "the most" into [", the, most, "], then the model will be quote boundary aware.

\section{Conclusion}
In this paper, I implemented a model with Bi-directional attention flow layer, connected with a Multi-layer LSTM encoder, connected with one start-index decoder and one conditioning end-index decoder and achieved great result. The best single model achieves an F1 score of 73.97\% and EM score of 64.95\% on test set. Based on the error analysis, 1) introducing number and special characters format feature  2) introducing part-of-speech tag  3) introducing name entity feature 4) improving tokenization logic will further improve the model performance. 

\subsubsection*{References}
\small{
[1] P. Rajpurkar, J. Zhang, K. Lopyrev, and P. Liang, Squad: 100, 000+ questions for machine comprehension of text. CoRR, abs/1606.05250, 2016.

[2] Minjoon Seo, Aniruddha Kembhavi, Ali Farhadi, and Hannaneh Hajishirzi. Bidirectional attention flow for machine comprehension. arXiv preprint arXiv:1611.01603, 2016.

[3] Shuohang Wangand JingJiang. Machine comprehension using match-lstm and answerpointer. arXiv preprint arXiv:1608.07905, 2016.

[4] Danqi Chen, Adam Fisch, Jason Weston, and Antoine Bordes. Reading wikipedia to answer open-domain questions. arXiv preprint arXiv:1704.00051, 2017.

[5] Jeffrey Pennington, Richard Socher, Christopher D. Manning. Glove: Global vectors
for word representation. 2014.

[6] Tensorflow org: https://www.tensorflow.org/
}

\end{document}